\documentclass[runningheads]{llncs}
\usepackage{graphicx}
\usepackage{times}
\usepackage{epsfig}
\usepackage{amsmath}
\usepackage{amssymb}
\usepackage{makecell} 
\usepackage{multirow} 
\usepackage{xcolor}
\usepackage{caption}
\usepackage{subcaption}
\usepackage{array}
\usepackage{url}
\usepackage{hyperref}
\usepackage[misc,geometry]{ifsym}

\hypersetup{
  colorlinks = true,
  linkcolor =red,
  citecolor = green,
  urlcolor = blue
}

\newcommand\blfootnote[1]{%
  \begingroup
  \renewcommand\thefootnote{}\footnote{#1}%
  \addtocounter{footnote}{-1}%
  \endgroup
}

\newcolumntype{?}[1]{!{\vrule width #1}}
\newcolumntype{L}[1]{>{\raggedright\arraybackslash}m{#1}}
\newcolumntype{C}[1]{>{\centering\arraybackslash}m{#1}}
\newcolumntype{R}[1]{>{\raggedleft\arraybackslash}m{#1}}
\begin{document}
\title{Context-Aware Chart Element Detection}
%
\titlerunning{CACHED: \textbf{C}ontext-\textbf{A}ware \textbf{Ch}art \textbf{E}lement \textbf{D}etection}
%
\author{Pengyu Yan$^*$ \and
Saleem Ahmed$^*$ \and
David Doermann}

\authorrunning{P. Yan et al.}

\institute{Department of Computer Science and Engineering,\\
University at Buffalo, Buffalo, NY \\
\email{\{pyan4, sahmed9, doermann\}@buffalo.edu} 
}
\maketitle              
\blfootnote{$^*$ Equal Contribution}

\begin{abstract}
As a prerequisite of chart data extraction, the accurate detection of chart basic elements is essential and mandatory. In contrast to object detection in the general image domain, chart element detection relies heavily on context information as charts are highly structured data visualization formats. To address this, we propose a novel method \textbf{CACHED}, which stands for \textbf{C}ontext-\textbf{A}ware \textbf{Ch}art \textbf{E}lement \textbf{D}etection, by integrating a local-global context fusion module consisting of visual context enhancement and positional context encoding with the Cascade R-CNN framework. To improve the generalization of our method for broader applicability, we refine the existing chart element categorization and standardized 18 classes for chart basic elements, excluding plot elements. Our CACHED method, with the updated category of chart elements, achieves state-of-the-art performance in our experiments, underscoring the importance of context in chart element detection. Extending our method to the bar plot detection task, we obtain the best result on the PMC test dataset. Our code and model are available at \url{https://github.com/pengyu965/ChartDete}.


\end{abstract}
\section{Introduction}
\label{introduction}
\begin{figure}[tb!]
    \centering
    \includegraphics[width=\textwidth]{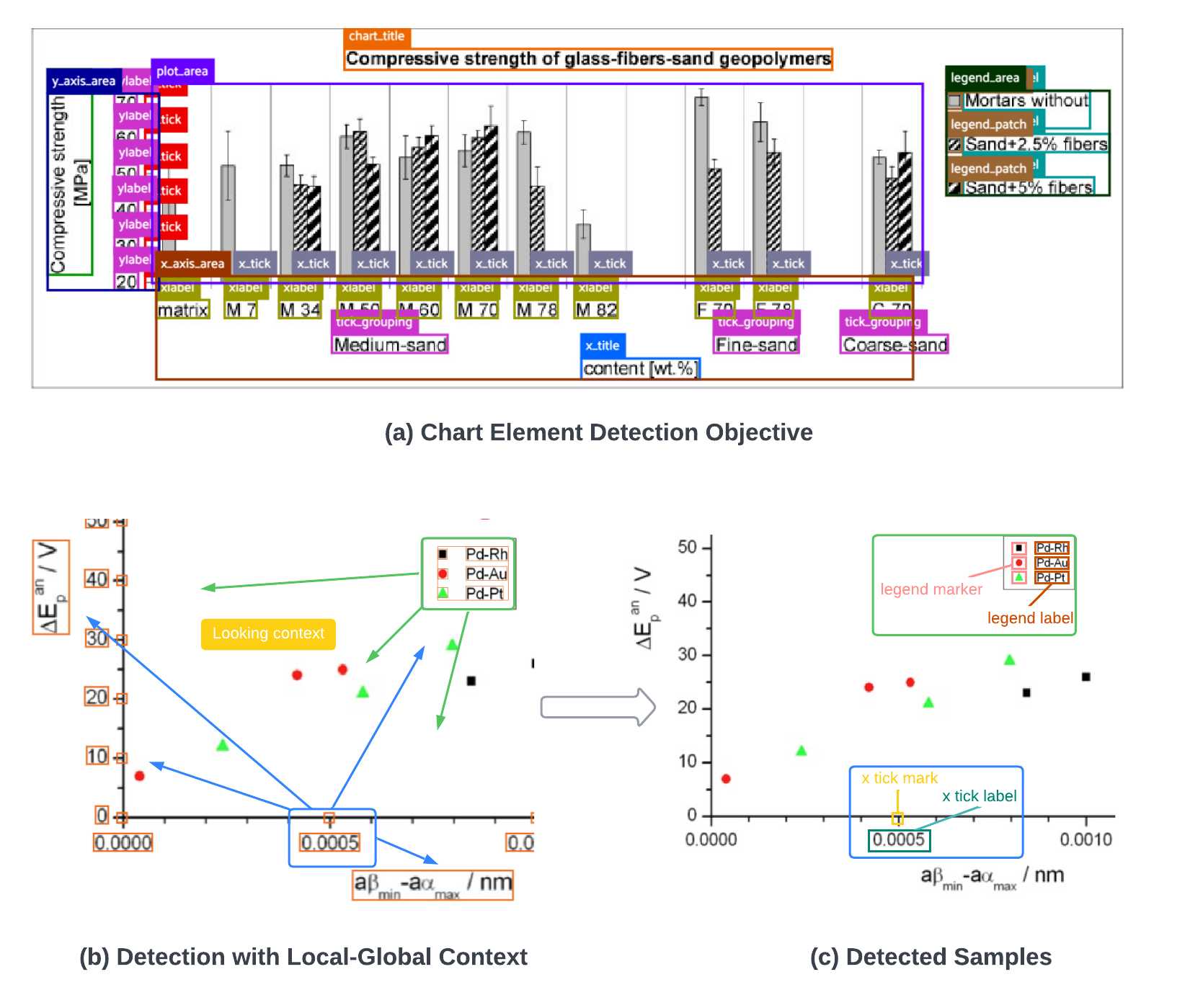}
     \caption{\textbf{Context-aware Chart Element Detection}. (a) is a sample of chart element detection objective, (b) and (c) illustrate the detection in chart images relies on context.}
    \label{chart_element_detection}
\end{figure}

Charts are highly abstract data visualization formats, which are convenient for readers to obtain trends or comparisons between different entities but hard to extract the exact data value from them. Therefore, automated chart data extraction could reduce the human effort to summarize data from scientific, financial analysis, marketing charts, etc. The chart data extraction typically involves but is not limited to, element detection, text OCR, data interpretation, and semantic data conversion. Basic element detection is the most fundamental part of chart data extraction and would affect all downstream tasks. Thus, accurately detecting and recognizing the basic elements of the chart outside the plot area (see Fig.~\ref{chart_element_detection}) is the first critical step. Basic element detection would be challenging due to the highly diverse chart design.\par 

Several methods~\cite{luo2021chartocr,ma2021towards} have been proposed for data extraction but only focused on data plot detection, such as bar plot and line key points, while neglecting most of the fundamental element detection. Meanwhile, some works~\cite{oglan2015chartsense,balaji2018charttext,hassan2023lineex} use often-seen standard two-stage detectors to detect chart elements. These two-stage detectors~\cite{rcnn,fast-rcnn,faster-rcnn,cascade-rcnn,mask-rcnn} have a limited ability to utilize the context in images, where context is vitally important for accurate detection in the chart images domain. Unlike common objects in general images, many elements in chart images share a similar visual appearance but different roles. It could only be distinguished by referring to the context, \emph{e.g.}, legend label and tick labels are text blocks with different roles according to the functioning position and relationship to the other elements in the chart images (as shown in Fig.~\ref{chart_element_detection}(b)(c)). Therefore, a detector that uses local-global context features is needed to tackle the chart element detection task. Additionally, a comprehensive and reasonable categorization of the chart elements would help improve generalization on various charts. \par 


In this paper, we propose a context-aware chart element detection method by integrating the local-global context fusion module between the cascaded RoI head in Cascade R-CNN to draw the importance of context in charts. As Fig.~\ref{framework} shows, the local-global context fusion module contains two parts--visual context enhancement (VCE) and positional context encoding (PCE). The VCE enables the model to gain better visual context information by incorporating the global feature map into the local feature map. The PCE allows the model to learn the particular object distribution pattern from the bbox coordinates. Besides, we look into multiple datasets for analyzing and refining chart element categorization. A total of 18 classes of chart elements are summarized, excluding any plot elements. The dataset is updated accordingly for model training and testing. The quantitative evaluation and qualitative analysis of samples show that our method achieves accurate chart element detection. \textbf{Overall, our contributions can be summarized as follows:} \par 

\begin{itemize}
    \item A detector with the local-global context fusion module is proposed for accurate chart element detection. The module emphasizes the context from two aspects - visual and positional features. Our method achieves state-of-the-art results on the chart element detection task.
    \item Refine chart element categorization and generate additional structural-area objects to assist the detector in better chart understanding. A total of 18 classes are summarized, and the accordingly updated PMC dataset can be accessed at {\small\url{https://github.com/pengyu965/ChartDete}}
    \item Our method and several common two-stage detectors, including those used in existing related works, are trained and evaluated on the updated datasets. This can offer an overview of the performance of these methods on such tasks. 
\end{itemize}

\section{Related Work}
\subsection{Object Detection}
Since 2014, many well-designed object detectors have been invented, and most of them could be divided into two kinds - one-stage detector~\cite{ssd,yolo,yolov2,yolov3} and two-stage detector~\cite{rcnn,fast-rcnn,faster-rcnn,cascade-rcnn,mask-rcnn}. They have a similar first part -- a convolutional neural network (CNN) based backbone like~\cite{he2016deep,xie2017aggregated,ghiasi2019fpn,howard2017mobilenets,zhang2018shufflenet,iandola2016squeezenet,sandler2018mobilenetv2,simonyan2014very} to extract visual features. The difference is two-stage detector involves a region proposal module to generate category-independent region proposals and then 
classify these objects and refine their localization in the second stage, while one-stage detectors directly predict the object position and category from input images and their feature maps. Intuitively second-stage detectors like Faster R-CNN~\cite{faster-rcnn} and Cascade R-CNN~\cite{cascade-rcnn} achieve higher accuracy in object localization and recognition but sacrifice the inference speed. The chart element detection task demands high accuracy on object localization and classification than inference speed, so we stay with the two-stage detectors for the task.\par 

\subsection{Transformers}
With the success of Transformer-based methods~\cite{transformer,bert} in natural language processing, many works~\cite{lxmert,visualbert,dvqa} put effort into adopting Transformer in vision-related tasks. Transformer could extract strong feature representations from both visual and text inputs. In 2020, Carion et al.~\cite{detr} proposed the first Transformer-based end-to-end object detector. However, DETR lacks training stability on small-scale datasets and cannot accurately localize and recognize small or overlapping objects. In a more recent work, \cite{swin-transformer} introduced a hierarchical Transformer that leverages shifted windows to extract features from input images, and the resulting framework serves as a new backbone for two-stage detectors. This new backbone has been shown to be effective in improving the performance of two-stage detectors.  Due to the attention mechanism in Transformer architecture, the Swin Transformer can draw spatial attention and obtain context-aware image features. The Swin-Transformer backbone is used exclusively in our model and experiments.\par 

\begin{figure}[tb!]
    \centering
    \includegraphics[width=0.9\textwidth]{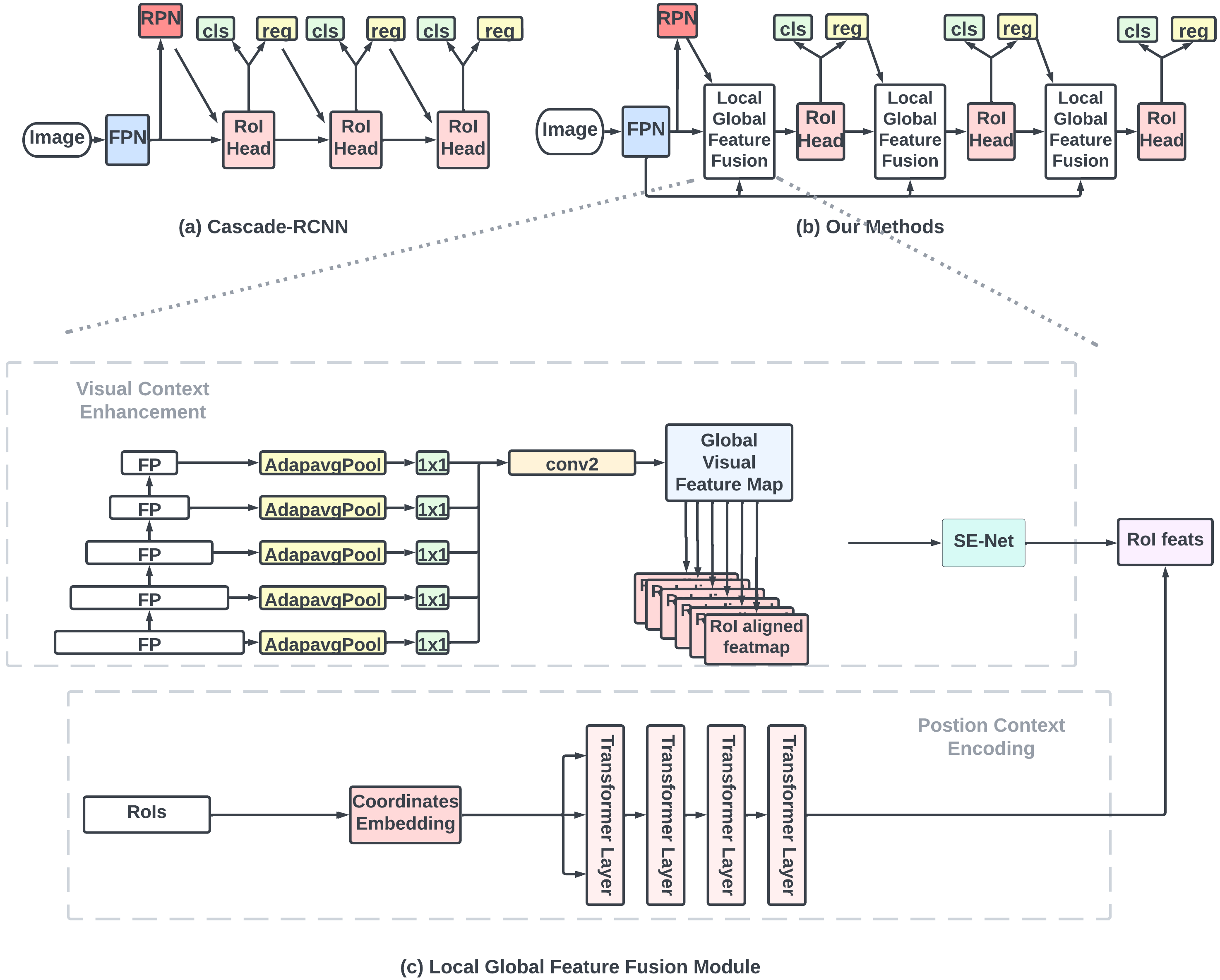}
    \caption{\textbf{Framework of our Method}. (a) is the standard Cascade R-CNN model, (b) is the overall framework of our method, and (c) shows the details of the local-global feature fusion module in our method.}
    \label{framework}
\end{figure}

\subsection{Chart Data Extraction}
In 2017, Jung et al.~\cite{oglan2015chartsense} proposed a semi-interactive system to extract underlying data from charts. However, it uses human interaction as the first step to set the starting and ending points. Then rule-based methods are utilized to interpret the data value. This method heavily relies on human interaction for data extraction and would fail in complicated cases. From 2019-2022, Davila et al. ~\cite{ICDAR2019ChartCompe,2020ChartCompetition,icpr2022ChartCompetition} organized the chart harvesting competition and offered two valuable datasets: Human-annotated real-world chart images from PubMed Central documents (PMC dataset) and Adobe synthetic chart dataset. As a participating group in the 2020 chart harvesting competition, Ma et al.~\cite{ma2021towards} use Cascade R-CNN and a heatmap-based keypoint detector to detect bar box and line key points, respectively. It focused on data plot detection within the plot area, while in data interpretation, they used the elements' ground truth with the predicted data plots for semantic conversion. Later, another EXCEL400K dataset is proposed in ~\cite{luo2021chartocr} by utilizing the Microsoft Excel API. Similarly to CornerNet~\cite{law2018cornernet}, an hourglass network backbone with key-point heatmap generation is used to predict and group the sets of corner points on objects. Using the key point detection method, the author generalized their detection on different charts. However, the dataset proposed in this work only has data plot annotations within the plot area, and most essential basic elements are left without annotations. \par

\section{Our Method}
In this section, we explain our CACHED method in the following aspects: the local-global context fusion module, which consists of visual context enhancement and positional context encoder; loss function for class-wise objects imbalance; the standardized categorization for broader applicability and generalization.\par 

\subsection{Local-Global Context Fusion Module}
As Fig.~\ref{framework} shows, local-global context fusion modules are designed and integrated between each RoI head for context extraction and fusion. It brings local-global visual features and relative positional features towards each region of interest before sending them to the RoI head for regression and classification. The three modules share the same architecture but are trained with unshared weights, and each module consists of the following two parts.\par 

\subsubsection{Visual Context Enhancement (VCE)}

Although the field of view of each anchor would increase with stacking of the convolutional neural network (CNN) layer, it is still limited to the local field, which has fragile context awareness. However, accurate element detection and labeling in chart images require a much larger field of context.(see Fig.~\ref{chart_element_detection}). For instance, a text block can be classified as a legend label by the context that nearly comes from the whole image, where the legend label is beside some (legend) markers that share the same color with plot elements in the plot area. To address this issue, similar to SCNet~\cite{han2017scnet}, we introduce visual context enhancement (VCE) by incorporating the feature maps from the backbone feature pyramid as global visual features and bringing them into each region of interest. These global visual features are combined with each RoI-aligned local feature map to amplify local-global visual context. As shown in Fig.~\ref{framework}(c), we first average pool the feature maps from all stages of the feature pyramid to the same size $[N, 256, 7, 7]$, then concatenate them together, where the shape becomes $[N, 256*5, 7, 7]$. A convolutional neural network further abstracts these global visual features and reduces the number of channels. Then the global feature maps are attached to each RoI-aligned feature map. The feature map for each region of interest with size $[N, 512, 7, 7]$ now consists of local visual features in the first half channels and global visual features in the rest of the channels. A SE-Net and a 1x1 convolutional layer are followed to fuse the local-global visual features channel-wise and reduce the number of channels, respectively. The final feature maps with size $[N, 256, 7, 7]$, which align with the original RoI feature map dimension but contain rich local-global visual context, are used as visual feature representations for each region of interest. \par


\subsubsection{Positional Context Encoding (PCE)}

In the general image domain, the detection of common objects such as people, vehicles, and animals depends mainly on visual features, and the positions of the objects in the image have less impact on accurate localization and classification. However, since chart images are highly structured data visualization formats, where object rendering always follows some specific patterns, \emph{e.g.}, x-tick labels are always below or beside the x-tick marks and chart titles typically appear on the edges of the image while rarely appearing in the center part, the relative positional context among the chart elements is very trivial. We propose a Transformer-based positional context encoder to obtain the positional context feature. As Fig.~\ref{framework} shows, the Transformer-based architecture draws attention among objects' bounding-box coordinates. Firstly, we normalize the 4-dim coordinates to $[0,1]$ by the height and width of the input images. Then a linear layer takes the normalized coordinates and embeds them into a 512-dim vector. Zero mask paddings are filled in the embedded bbox vector sequence, where the length of the final embedding bbox sequence is fixed to the max block size of 1024. The block size of 1024 is enough to cover the maximum sampling bbox number from the RPN results in training and testing, which is set to a maximum of 1000. We use each head's output to represent the relative positional context information from the corresponding bbox to all other bboxes. Then, the encoding vector for each bbox is concatenated to its RoI visual feature vectors. \par 


\begin{figure}[tb]
    \centering
    \includegraphics[width=0.7\textwidth]{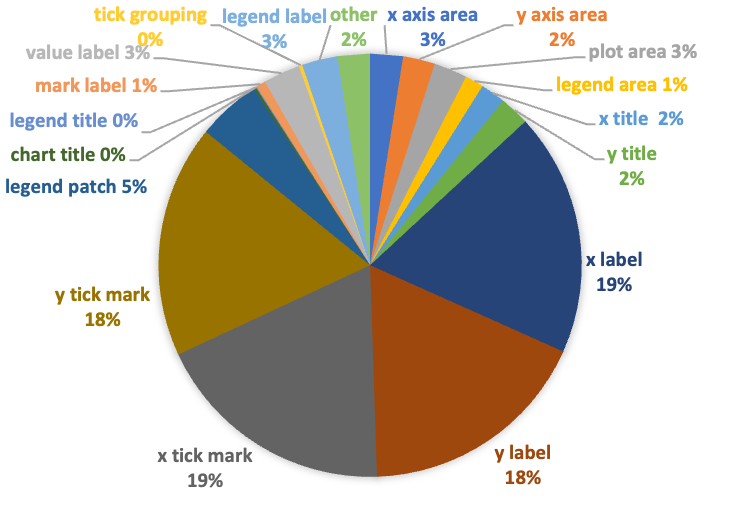}
    \caption{\textbf{The class-wise chart object number distribution in PMC dataset}. Chart images have naturally unbalanced objects class-wisely. The number of x/y tick labels and x/y tick marks is inevitably larger than others. Some elements are extremely small, like tick groupings and legend titles.}
    \label{category_statistic}
\end{figure}

\subsection{Loss for class-wise objects imbalance}
Unlike the general image domain data imbalance can be reduced by augmentation, the unbalanced number of objects among different categories is the natural effect in chart images and is hard to undermine, \emph{e.g.}, the number of tick marks and tick labels would always be multi-times larger than the number of chart titles and legend labels (see Fig.~\ref{category_statistic}). We use the Focal Loss~\cite{lin2017focal} on the classification to undermine the object imbalance in chart images:
\begin{equation}
    L_{cls} = -\sum_{i=1}^{i=n}(i-p_i)^\gamma \log_b(p_i) 
\end{equation}\par 

Then, smooth L1 loss is used for bounding box regression and balancing it with classification:

\begin{equation}
    loss_{\mu} = L_{cls}(\mu) + \lambda[\mu\geq1]L_{loc}(t^\mu,v)
\end{equation}
where $L_{loc}$ is the localization loss between the regression results for class $\mu$ and the regression targets. $\lambda$ is used to tune the loss for outliers where their loss is greater than 1.\par

\subsection{Categorization Refinement}
The categorization of chart elements in the dataset is crucial for detection performance. We analyze the datasets with the annotations of the chart elements and comprehend the categorization of the elements to generalize the various chart situations (see Table~\ref{adopted_category}). \par 

\begin{figure}[!htb]
    \centering
    \includegraphics[width=\textwidth]{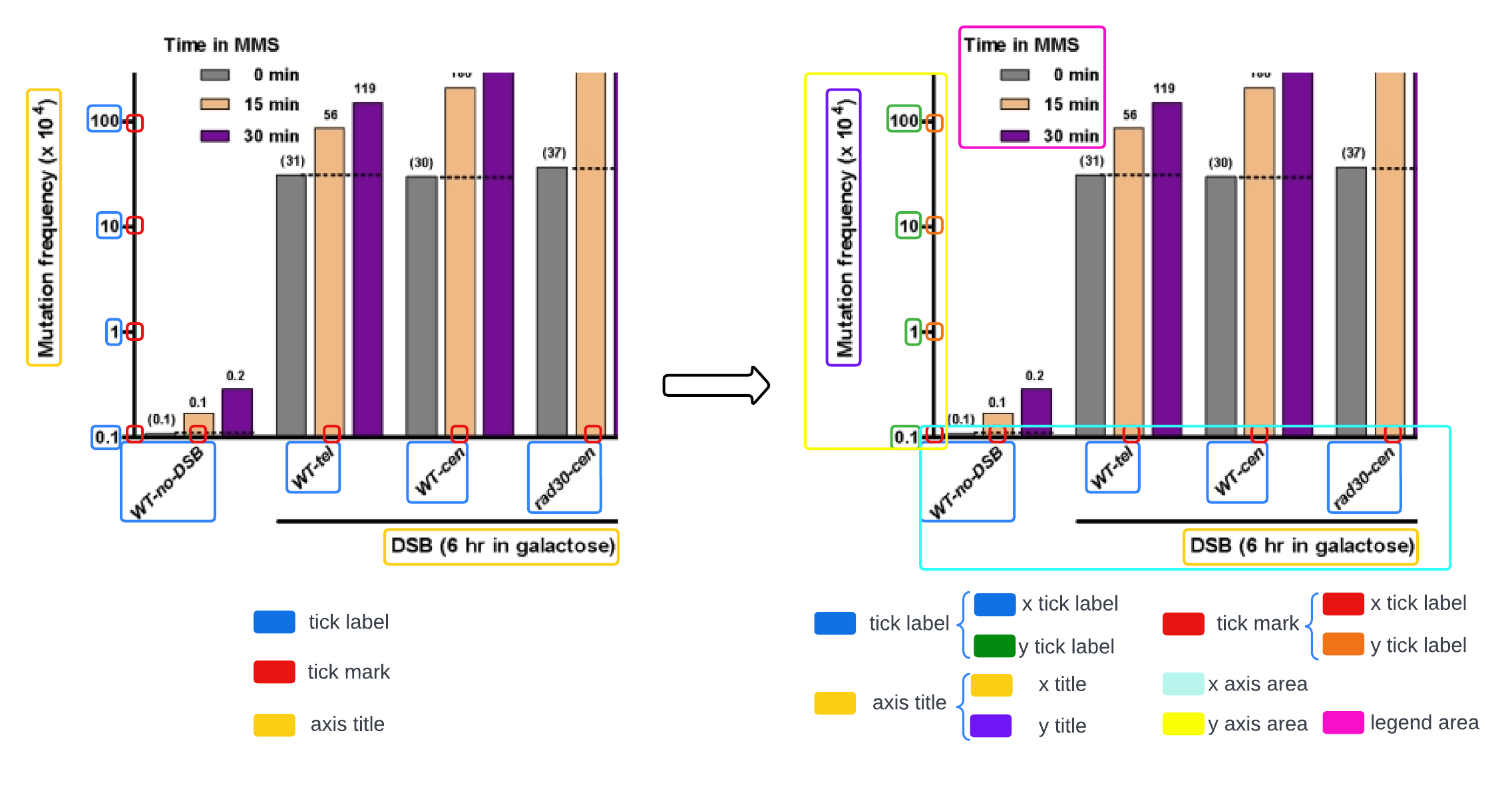}
    \caption{\textbf{Categorization Refinement}. It only shows the updated categories for illustration. Some elements are not visualized.} 
    \label{categorization_refinement}
\end{figure}

From the chart competition~\cite{2020ChartCompetition,icpr2022ChartCompetition}, the Adobe Synthetic and PMC datasets offer the most valuable annotations for chart element detection tasks (a detailed introduction of the datasets can be found in the experiment section). We refer to these two datasets for element categorization. In Table~\ref{adopted_category}, the second column is the category in the Adobe Synthetic dataset, and the third column is the category annotated in the PMC dataset. In the PMC dataset, all objects related to the axes, such as the axis title, the tick label, and the tick mark, are jointly labeled without separation into the x and y axes. Early experiments showed that if we jointly treat the items based on the x- and y-axis as the same category, it would be easier to separate these labels later by post-processing. This is caused by the definition of the x-axis and y-axis, in which the x-axis is defined as the axis with independent value/label, while the y-axis has dependent values. Such definitions are more based on the content's semantic meaning than on the visual or positional information. Therefore, separating these elements by axis in advance would help avoid this issue when using the trained model for the application. Meanwhile, these separated labels also assist our method in understanding the chart's content and potentially obtaining better detection results. \par

\begin{table}[htb]
    \centering
    \caption{\textbf{Categorization Refinement}. We look into the Adobe Synthetic Dataset and PMC Dataset from Chart Competition~\cite{2020ChartCompetition,icpr2022ChartCompetition} and comprehend the element categories shown in the third column.}\vspace{0.5em}
    \begin{tabular}{|C{8em}|C{8em}|C{8em}|C{10em}|}
        \Xhline{1.5pt}
         & \textbf{Synthetic} & \textbf{PMC} & \textbf{Refined Categories}  \\
        \Xhline{1.5pt}
        
        \multirow{14}{*}{\makecell[c]{\textbf{Chart Basic}\\\textbf{Skeleton Elements}}} & x-axis title & \multirow{2}{*}{axis title} & x-axis title \\
        \cline{2-2} \cline{4-4}
        
         & y-axis title &  & y-axis title \\ 
        \cline{2-4} 
        
         & x tick label & \multirow{2}{*}{tick label} & x tick label \\
        \cline{2-2} \cline{4-4}
        
         & y tick label & & y tick label \\
        \cline{2-4} 
        
         & x tick mark & \multirow{2}{*}{tick mark} & x tick mark \\
        \cline{2-2} \cline{4-4}
        
         & y tick mark & & y tick mark \\
        \cline{2-4} 
        
         & chart title & chart title & chart title \\ 
        \cline{2-4} 
        
         & legend patch & legend marker & legend marker \\
        \cline{2-4} 
        
         & legend label & legend label & legend label \\ 
        \cline{2-4} 
        
         & - & legend title & legend title \\
        \cline{2-4}
        
         & \multirow{2}{*}{plot text} & value label & value label \\
        \cline{3-4} 
        
         &  & mark label & mark label \\
        \cline{2-4}  
        
         & - & tick grouping & tick grouping \\
        \cline{2-4} 
        
         & - & others & others \\
        \Xhline{1pt}
        \multirow{4}{*}{\makecell[c]{\textbf{Chart}\\\textbf{Structural Area}}} & plot area & plot area & plot area \\
        
         & \multirow{3}{*}{-} & \multirow{3}{*}{-} & x-axis area \\
         
         & & & y-axis area \\ 
         
         & & & legend area \\ 
    \Xhline{1.5pt}
    \end{tabular}
    \label{adopted_category}
\end{table}

The categories in the second part of Table~\ref{adopted_category} are four additional labels of structural objects that cover the specific area in the chart images. We break down the chart images into the four most crucial structures --- the x-axis area, the y-axis area, the plot area, and the legend area. The definitions of these four areas are as follows:\par

\begin{itemize}
    \item \textbf{Plot area}: The plotting area is formed by the x and y axes.
    \item \textbf{X/Y axis area}: Cover all x/y ticks, x/y labels, and x/y-axis titles.
    \item \textbf{Legend area}: The area covers all items related to the legend, including legend labels, legend markers, and legend titles.
\end{itemize} 

A sample of category refinement is shown in Fig.~\ref{categorization_refinement}. We can see that each axis-related element is separated and structural-area objects cover the specific area in the chart image. After refinement, 18 categories for the chart elements are summarized. Based on the new categories, we update the PMC dataset accordingly by employing rule-based methods to separate the axis-related elements and generate three additional structure-area elements than the originally offered plot area. Then the annotation format is converted to the standard COCO object detection format for convenience. The updated PMC dataset with conversion tools can be accessed from the link in the contribution summary from Section~\ref{introduction}.\par 





\section{Datasets and Experiments}
In this section, we introduce the existing datasets used in our experiments and perform quantitative evaluation and qualitative analyzes. \par 

\subsection{Dataset}

\subsubsection{Adobe Synthetic Dataset.} The adobe synthetic dataset was first proposed in 2019~\cite{ICDAR2019ChartCompe} and refined in 2020 Chart-infographic Competition~\cite{2020ChartCompetition}. This dataset is synthesized using Matplotlib and contains 14400 images for 12 types of charts. The annotations include the chart data information and all elements' label and location. Such annotations are valuable for chart classification, element detection, and data extraction tasks. However, the diversity of chart samples in this dataset is severely limited. The best results showed in ICPR 2020 Chart Competition~\cite{2020ChartCompetition} in chart classification, text role classification, tick mark label association, and legend marker detection are close to 100\%, further indicating the limited variance of the data. Low data variance typically causes unstable performance and low generalization from the trained deep-learning model on samples outside the dataset. We only refer to the annotation standards in this dataset for element categorization refinement. \par 

\subsubsection{PubMed Central (PMC) Chart Dataset.} This dataset was released and updated with the Chart Competition in ICDAR 2019~\cite{2020ChartCompetition}, ICPR 2020~\cite{icpr2022ChartCompetition} and ICPR 2022~\cite{icpr2022ChartCompetition}. Unlike the Adobe Synthetic dataset, the PMC dataset is a real-world dataset collected from PubMed Central Documents and manually annotated. Taking into account the much more diverse and high-fidelity samples in this dataset, the PMC dataset became the primary dataset in most recent chart competitions. The most up-to-date PMC dataset is released in the ICPR 2022 CHART-Infographic competition~\cite{icpr2022ChartCompetition}, which contains 5614 images for chart element detection, 4293 images for final plot detection and data extraction, and 22924 images for chart classification. Although slightly limited by the number of available training samples due to the time-consuming human annotation process, this real-world dataset is most valuable, and much more challenging than most synthetic datasets. \par 

\subsubsection{Excel400K} In~\cite{luo2021chartocr}, the author proposed the ExcelChart400k dataset, which contains a total of 360k training samples. This dataset is generated using the Excel data sheet and was used to train the plot element detection model in~\cite{luo2021chartocr}. However, the dataset is focused only on annotating data plots inside the plot area, and most basic elements are left without annotation. We use this additional dataset to qualitatively evaluate the results of our method.\par

After conducting early experiments, we observed that including the Adobe synthetic training dataset decreased the detection performance on real-world chart images. This was attributed to the uniform chart rendering patterns and limited sample variance in the synthetic dataset. Taking into account the high diversity and real-world chart data distribution, we use the PMC dataset as our primary dataset for training and quantitative evaluation purposes.\par 



\begin{table}[tb]
    \centering
    \small
    \caption{\textbf{Comparison of Methods in ICPR 2022 Chart Competition.} Evaluation on task 2 and task 3 in chart competition~\cite{icpr2022ChartCompetition} is based on the original PMC categorization. We train our method with the updated categories but rewind the predicted results backward to be compatible with the chart competition evaluation metrics.}\vspace{0.5em}
    \begin{tabular}{r ?{1pt} c || c | c | c}
        \Xhline{1.5pt}
        \multirow{2}{*}{Team Methods} & \textbf{Task 2 Text Detection} & \multicolumn{3}{c}{\textbf{Task 3 Text Role Classification}} \\
        \cline{2-5}
         & Average IoU  & \hspace{0.3em} Recall \hspace{0.3em} & Precision & F-measure  \\
        \Xhline{1.5pt}
        six\_seven\_four & 0.435 & - & - & -  \\
        IIIT\_CVIT\_Chart\_Understanding & 0.790 & - & - & \textbf{0.821} \\
        Ystar & 0.810 & - & - & - \\
        UB-ChartAnalysis & 0.820 & - & - & 0.736 \\
        \textbf{Ours} & \textbf{0.869} & \textbf{0.735} & \textbf{0.846} & \textbf{0.787} \\
        \Xhline{1.5pt}
    \end{tabular}
    \label{legacy_result_comparison}
\end{table}


\subsection{Quantitative Evaluation}

To quantitatively evaluate our method, we have undertaken three experiments: (i) ICPR 2022 Chart Competition Evaluation~\cite{icpr2022ChartCompetition}, wherein we employed backward adapted results from our method; (ii) COCO object detection evaluation with the refined PMC dataset; and (iii) an extended experiment on detecting bar plots.

\subsubsection{ICPR 2022 Chart Competition Evaluation} 

We compare with the results from the ICPR 2022 Chart Competition~\cite{icpr2022ChartCompetition}. However, our method shares different detection routines, where the chart competition treats the text block detection task as a 1-class detection task (task 2) and leaves the recognition as an additional task (task 3). In the chart competition, teams could use the ground truth of previous tasks for the current task. Our method follows the regular detection routine (localization + recognition). After refining the categorization, the overall categories (see Table~\ref{adopted_category}) expand from the original PMC dataset used in the competition. Taking these two facts, we compare by adapting our prediction results backwards to be compatible for evaluation and splitting the prediction result into task 2 and task 3. Table~\ref{legacy_result_comparison} shows the result of the official 2022 chart competition. Our method in task 2 outperforms the best result from `UB-ChartAnalysis'. In task 3, our method detects and recognizes all elements once, without taking the element localization ground truth as prior knowledge, and this may result in a lower recall for task 3 due to imperfect detection. Overall, we achieve the best detection result on task 2 and the comparable F-measure score on task 3. \par 

\begin{table}[tb]
    \centering
    \caption{\textbf{COCO Evaluation on Refined PMC Dataset.} First 4 rows show our trained public detector models, and the last 3 rows are our method ablation study.} \vspace{0.5em}
    \begin{tabular}{l c c c c c c c c}
        \Xhline{0.8pt}
        \textbf{Model} &\textbf{Context Module} & \textbf{Backbone} & $AP$ & $AP_{50}$ & $AP_{75}$ & $AP_{S}$ & $AP_{M}$ & $AP_{L}$ \\
        \Xhline{0.5pt}
        DETR & -  & - & 0.536 & 0.762 & 0.572 & 0.384 & 0.555 & 0.809 \\
        Faster R-CNN & - & ResNeXt101-FPN & 0.665 & 0.815 & 0.737 & 0.557 & 0.697 & 0.827 \\
        Cascade R-CNN & - & ResNeXt101-FPN & 0.696 & 0.825 & 0.759 & 0.579 & 0.725 & 0.899 \\
        Cascade R-CNN & - & SwinT-FPN & 0.699 & 0.838 & 0.772 & 0.589 & 0.732 & 0.885 \\
        \hline 
        \hline
        Cascade R-CNN & \textbf{PCE} & SwinT-FPN & 0.708 & 0.842 & 0.775 & 0.591 & 0.742 & 0.903 \\ 
        Cascade R-CNN & \textbf{VCE+PCE} & SwinT-FPN & 0.713 & \textbf{0.851} & 0.786 & 0.597 & 0.741 & 0.909 \\
        Cascade R-CNN$^L$ & \textbf{VCE+PCE} & SwinT-FPN & \textbf{0.729} & 0.845 & \textbf{0.790} &\textbf{0.602} & \textbf{0.763} & \textbf{0.939} \\
        \Xhline{0.8pt}
    \end{tabular}
    \label{results_comparison}
\end{table}

\subsubsection{COCO Evaluation on Refined PMC Datasets}

To provide a better general overview of the performance of our method, we use the COCO object detection evaluation metric~\cite{lin2014microsoft_coco} to evaluate our method with several often used two-stage detectors (shown in Table~\ref{results_comparison}). To our best knowledge, there isn't any trained chart element detection model conducted on PMC datasets publicly available for inference and fine-tuning. We trained these two-stage detectors on the refined PMC dataset at our end. In Table~\ref{results_comparison}, the first part includes results from popular two-stage detectors. DETR converges slowly in training due to the scale of our dataset and the Transformer architecture. The accuracy of small object detection from DETR also lags behind. An enormous amount of small elements in chart images result in limited overall accuracy. The Cascade R-CNN with Swin-Transformer performs best in the standard Cascade R-CNN model zoo. The second part shows our methods with several ablated setups on the local-global context module, which consists of visual context enhancement (VCE) and positional context encoding (PCE). Since the Swin-Transformer backbone can draw spatial attention and potentially obtain context-aware visual feature representations, we made an ablation by only integrating the positional context encoding (PCE) into the Swin-Transformer-based Cascade R-CNN. As expected, the Swin-Transformer-based Cascade R-CNN with the PCE performs better than the standard Swin-Transformer Cascade R-CNN. Although the Swin-Transformer can obtain context-aware visual features, adding VCE could further enhance the context extraction capability, giving better results. The focal loss with balanced smooth L1 loss can help with sample imbalance problems, and our method with this loss setup achieves state-of-the-art performance (see the last row in Table~\ref{results_comparison}). \par

\begin{table}[tb]
    \centering
    \caption{\textbf{Bar detection evaluation on PMC test dataset.} The comparison results are from~\cite{ma2021towards}, and we evaluate our method with the same evaluation metric. The first three columns are F-measure with different IoU thresholds, and the last column is the calculated score using Chart Competition~\cite{2020ChartCompetition,icpr2022ChartCompetition} evaluation criteria.} \vspace{0.5em}
    \begin{tabular}{L{12em} C{6em} C{6em} C{6em} C{6em}}
    \Xhline{1pt}
        Model & IoU=0.5 & IoU=0.7 & IoU=0.9 & Score\_a \\
    \Xhline{1pt}
        SSD & 43.65 & 26.28 & 2.67 & 25.83 \\

        YOLO-v3 & 58.84 & 36.14 & 4.14 & 60.97 \\

        Faster R-CNN & 66.37 & 60.88 & 29.13 & 70.03 \\

        Faster R-CNN+FPN & 85.81 & 78.05 & 31.30 & 89.65 \\

        Cascade R-CNN+FPN & 86.92 & 83.53 & 55.32 & 91.76 \\

        \textbf{Our Methods} & \textbf{89.30} & \textbf{88.73} & \textbf{76.94} & \textbf{93.75} \\ 
    \Xhline{1pt}
    \end{tabular}
    \label{bar_eval}
\end{table}

\subsubsection{Extended Experiment on Bar Detection}

Although our goal is to offer a robust basic element detection method, we extend our experiment to bar plot detection. We fine-tune our method using the PMC bar chart subset and test on the PMC test set (see Table~\ref{bar_eval}). The results of the first five models were from~\cite{ma2021towards}, and we evaluate our results with the same criteria. The first three columns are F-measure scores with different IoU thresholds - 0.5, 0.7, and 0.9. The last column, `Score\_a', is calculated using the ICPR 2022 chart competition~\cite{2020ChartCompetition} evaluation metric for task 6a. Our method achieves state-of-the-art performance on bar chart detection on the PMC dataset. \par

\begin{figure}
    \centering
    \includegraphics[width=\textwidth]{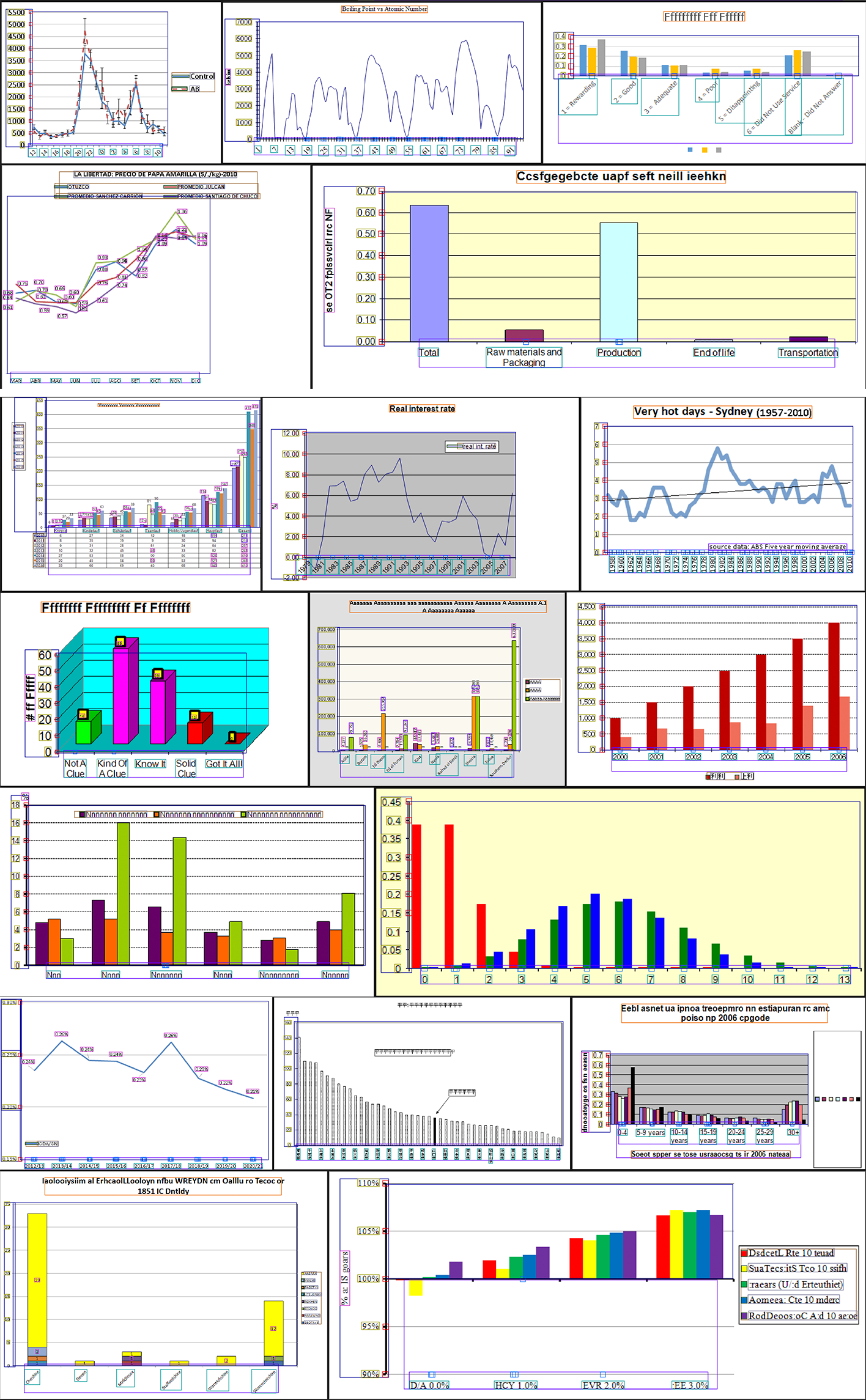}
    \caption{Visualization of Our methods on Excel400K.}
    \label{excel400k_vis}
\end{figure}

\subsection{Qualitative Evaluation on Element Detection}
Although the Excel400K dataset does not have the ground truth for chart basic elements, we visualize the prediction results from our method (see Fig.~\ref{excel400k_vis}) for qualitative evaluation. Our method is able to locate each element accurately on most samples in Excel400K datasets. However, our method struggles with the first sample in the third row of Fig.~\ref{excel400k_vis} as the table attached at the bottom confuses the detector due to the lack of similar samples in the PMC training dataset. Our method generally delivers accurate localization and classification of the basic elements in charts. \par

\section{Conclusion and Future Work}
In this work, we propose a method that focuses on the importance of visual and positional context in chart images for accurate chart element detection. The categories of chart elements are analyzed and refined to provide a better generalization of various chart designs, which could benefit data interpretation-related downstream tasks. Our method trained on the refined PMC dataset achieves state-of-the-art performance on the chart element detection task. \par 

Considering the context information, the text contains rich information. Due to the challenging OCR task on chart images, where many symbols are easily confused with characters or numbers, and the rotation is hard to detect when the text is short, we don't include the text embedding into the context extraction. In the future, robust OCR for extracting chart text and adding chart text embedding as additional context information to each region of interest may improve performance.\par

\newpage
\bibliographystyle{splncs04}
\bibliography{references}

\begin{thebibliography}{10}
\providecommand{\url}[1]{\texttt{#1}}
\providecommand{\urlprefix}{URL }
\providecommand{\doi}[1]{https://doi.org/#1}

\bibitem{balaji2018charttext}
Balaji, A., Ramanathan, T., Sonathi, V.: Chart-text: A fully automated chart
  image descriptor. arXiv preprint arXiv:1812.10636  (2018)

\bibitem{cascade-rcnn}
Cai, Z., Vasconcelos, N.: Cascade r-cnn: Delving into high quality object
  detection. In: Proceedings of the IEEE conference on computer vision and
  pattern recognition. pp. 6154--6162 (2018)

\bibitem{detr}
Carion, N., Massa, F., Synnaeve, G., Usunier, N., Kirillov, A., Zagoruyko, S.:
  End-to-end object detection with transformers. In: European Conference on
  Computer Vision. pp. 213--229. Springer (2020)

\bibitem{ICDAR2019ChartCompe}
Davila, K., Kota, B.U., Setlur, S., Govindaraju, V., Tensmeyer, C., Shekhar,
  S., Chaudhry, R.: Icdar 2019 competition on harvesting raw tables from
  infographics (chart-infographics). In: 2019 International Conference on
  Document Analysis and Recognition (ICDAR). pp. 1594--1599 (2019).
  \doi{10.1109/ICDAR.2019.00203}

\bibitem{2020ChartCompetition}
Davila, K., Tensmeyer, C., Shekhar, S., Singh, H., Setlur, S., Govindaraju, V.:
  Icpr 2020-competition on harvesting raw tables from infographics. In:
  International Conference on Pattern Recognition. pp. 361--380. Springer
  (2021)

\bibitem{icpr2022ChartCompetition}
Davila, K., Xu, F., Ahmed, S., Mendoza, D.A., Setlur, S., Govindaraju, V.: Icpr
  2022: Challenge on harvesting raw tables from infographics
  (chart-infographics). In: 2022 26th International Conference on Pattern
  Recognition (ICPR). pp. 4995--5001. IEEE (2022)

\bibitem{bert}
Devlin, J., Chang, M.W., Lee, K., Toutanova, K.: Bert: Pre-training of deep
  bidirectional transformers for language understanding. arXiv preprint
  arXiv:1810.04805  (2018)

\bibitem{ghiasi2019fpn}
Ghiasi, G., Lin, T.Y., Le, Q.V.: Nas-fpn: Learning scalable feature pyramid
  architecture for object detection. In: Proceedings of the IEEE/CVF Conference
  on Computer Vision and Pattern Recognition. pp. 7036--7045 (2019)

\bibitem{fast-rcnn}
Girshick, R.: Fast r-cnn. In: Proceedings of the IEEE international conference
  on computer vision. pp. 1440--1448 (2015)

\bibitem{rcnn}
Girshick, R., Donahue, J., Darrell, T., Malik, J.: Rich feature hierarchies for
  accurate object detection and semantic segmentation. In: Proceedings of the
  IEEE conference on computer vision and pattern recognition. pp. 580--587
  (2014)

\bibitem{han2017scnet}
Han, K., Rezende, R.S., Ham, B., Wong, K.Y.K., Cho, M., Schmid, C., Ponce, J.:
  Scnet: Learning semantic correspondence. In: Proceedings of the IEEE
  international conference on computer vision. pp. 1831--1840 (2017)

\bibitem{hassan2023lineex}
Hassan, M.Y., Singh, M., et~al.: Lineex: Data extraction from scientific line
  charts. In: Proceedings of the IEEE/CVF Winter Conference on Applications of
  Computer Vision. pp. 6213--6221 (2023)

\bibitem{mask-rcnn}
He, K., Gkioxari, G., Doll{\'a}r, P., Girshick, R.: Mask r-cnn. In: Proceedings
  of the IEEE international conference on computer vision. pp. 2961--2969
  (2017)

\bibitem{he2016deep}
He, K., Zhang, X., Ren, S., Sun, J.: Deep residual learning for image
  recognition. In: Proceedings of the IEEE conference on computer vision and
  pattern recognition. pp. 770--778 (2016)

\bibitem{howard2017mobilenets}
Howard, A.G., Zhu, M., Chen, B., Kalenichenko, D., Wang, W., Weyand, T.,
  Andreetto, M., Adam, H.: Mobilenets: Efficient convolutional neural networks
  for mobile vision applications. arXiv preprint arXiv:1704.04861  (2017)

\bibitem{iandola2016squeezenet}
Iandola, F.N., Han, S., Moskewicz, M.W., Ashraf, K., Dally, W.J., Keutzer, K.:
  Squeezenet: Alexnet-level accuracy with 50x fewer parameters and< 0.5 mb
  model size. arXiv preprint arXiv:1602.07360  (2016)

\bibitem{dvqa}
Kafle, K., Price, B., Cohen, S., Kanan, C.: Dvqa: Understanding data
  visualizations via question answering. In: Proceedings of the IEEE conference
  on computer vision and pattern recognition. pp. 5648--5656 (2018)

\bibitem{law2018cornernet}
Law, H., Deng, J.: Cornernet: Detecting objects as paired keypoints. In:
  Proceedings of the European conference on computer vision (ECCV). pp.
  734--750 (2018)

\bibitem{visualbert}
Li, L.H., Yatskar, M., Yin, D., Hsieh, C.J., Chang, K.W.: Visualbert: A simple
  and performant baseline for vision and language. arXiv preprint
  arXiv:1908.03557  (2019)

\bibitem{lin2017focal}
Lin, T.Y., Goyal, P., Girshick, R., He, K., Doll{\'a}r, P.: Focal loss for
  dense object detection. In: Proceedings of the IEEE international conference
  on computer vision. pp. 2980--2988 (2017)

\bibitem{lin2014microsoft_coco}
Lin, T.Y., Maire, M., Belongie, S., Hays, J., Perona, P., Ramanan, D.,
  Doll{\'a}r, P., Zitnick, C.L.: Microsoft coco: Common objects in context. In:
  Computer Vision--ECCV 2014: 13th European Conference, Zurich, Switzerland,
  September 6-12, 2014, Proceedings, Part V 13. pp. 740--755. Springer (2014)

\bibitem{ssd}
Liu, W., Anguelov, D., Erhan, D., Szegedy, C., Reed, S., Fu, C.Y., Berg, A.C.:
  Ssd: Single shot multibox detector. In: European conference on computer
  vision. pp. 21--37. Springer (2016)

\bibitem{swin-transformer}
Liu, Z., Lin, Y., Cao, Y., Hu, H., Wei, Y., Zhang, Z., Lin, S., Guo, B.: Swin
  transformer: Hierarchical vision transformer using shifted windows. arXiv
  preprint arXiv:2103.14030  (2021)

\bibitem{luo2021chartocr}
Luo, J., Li, Z., Wang, J., Lin, C.Y.: Chartocr: data extraction from charts
  images via a deep hybrid framework. In: Proceedings of the IEEE/CVF winter
  conference on applications of computer vision. pp. 1917--1925 (2021)

\bibitem{ma2021towards}
Ma, W., Zhang, H., Yan, S., Yao, G., Huang, Y., Li, H., Wu, Y., Jin, L.:
  Towards an efficient framework for data extraction from chart images. arXiv
  preprint arXiv:2105.02039  (2021)

\bibitem{oglan2015chartsense}
Oglan, V.A.: Chart sense: Common sense charts to teach 3-8 informational text
  and literature. Language Arts  \textbf{92}(5), ~368 (2015)

\bibitem{yolo}
Redmon, J., Divvala, S., Girshick, R., Farhadi, A.: You only look once:
  Unified, real-time object detection. In: Proceedings of the IEEE conference
  on computer vision and pattern recognition. pp. 779--788 (2016)

\bibitem{yolov2}
Redmon, J., Farhadi, A.: Yolo9000: better, faster, stronger. In: Proceedings of
  the IEEE conference on computer vision and pattern recognition. pp.
  7263--7271 (2017)

\bibitem{yolov3}
Redmon, J., Farhadi, A.: Yolov3: An incremental improvement. arXiv preprint
  arXiv:1804.02767  (2018)

\bibitem{faster-rcnn}
Ren, S., He, K., Girshick, R., Sun, J.: Faster r-cnn: Towards real-time object
  detection with region proposal networks. Advances in neural information
  processing systems  \textbf{28},  91--99 (2015)

\bibitem{sandler2018mobilenetv2}
Sandler, M., Howard, A., Zhu, M., Zhmoginov, A., Chen, L.C.: Mobilenetv2:
  Inverted residuals and linear bottlenecks. In: Proceedings of the IEEE
  conference on computer vision and pattern recognition. pp. 4510--4520 (2018)

\bibitem{simonyan2014very}
Simonyan, K., Zisserman, A.: Very deep convolutional networks for large-scale
  image recognition. arXiv preprint arXiv:1409.1556  (2014)

\bibitem{lxmert}
Tan, H., Bansal, M.: Lxmert: Learning cross-modality encoder representations
  from transformers. arXiv preprint arXiv:1908.07490  (2019)

\bibitem{transformer}
Vaswani, A., Shazeer, N., Parmar, N., Uszkoreit, J., Jones, L., Gomez, A.N.,
  Kaiser, {\L}., Polosukhin, I.: Attention is all you need. In: Advances in
  neural information processing systems. pp. 5998--6008 (2017)

\bibitem{xie2017aggregated}
Xie, S., Girshick, R., Doll{\'a}r, P., Tu, Z., He, K.: Aggregated residual
  transformations for deep neural networks. In: Proceedings of the IEEE
  conference on computer vision and pattern recognition. pp. 1492--1500 (2017)

\bibitem{zhang2018shufflenet}
Zhang, X., Zhou, X., Lin, M., Sun, J.: Shufflenet: An extremely efficient
  convolutional neural network for mobile devices. In: Proceedings of the IEEE
  conference on computer vision and pattern recognition. pp. 6848--6856 (2018)

\end{thebibliography}

\end{document}